%% file: ACMTargetAd2016.tex
\documentclass{sig-alternate-05-2015}
\usepackage{subfigure}

\makeatletter
\newif\if@restonecol
\makeatother

\usepackage[ruled,vlined,linesnumbered,commentsnumbered]{algorithm2e}

\usepackage{url}
\usepackage{verbatim}

\usepackage{graphicx}
\usepackage{amsmath}
\usepackage{algorithmic}

\usepackage{subfigure}

\usepackage{color}
\definecolor{gray}{gray}{0.95}
\usepackage{listings}
\usepackage{url}

\newcommand{\var}[1]{{\operatorname{\mathit{#1}}}}

\newcounter{def_count}
\newcounter{theorem}
\begin{document}






%

\permission{Permission to make digital or hard copies of all or part of this work for personal or classroom use is granted without fee provided that copies are not made or distributed for profit or commercial advantage and that copies bear this notice and the full citation on the first page. To copy otherwise, or republish, to post on servers or to redistribute to lists, requires prior specific permission and/or a fee.}
\conferenceinfo{TargetAd 2016: 2nd International Workshop on Ad Targeting at Scale,}{WSDM, February 22--25, 2016, San Francisco, CA, USA.}
\copyrightetc{{\em TargetAd 2016: 2nd International Workshop on Ad Targeting at Scale, WSDM, February 22--25, 2016, San Francisco, CA, USA.}}
\crdata{}

\title{Finding Needle in a Million Metrics: Anomaly Detection in a Large-scale Computational Advertising Platform}
%
%
%
%
%

\numberofauthors{2} 
%
\author{
%
%
\alignauthor
Bowen Zhou\\
       \affaddr{Turn Inc.}\\
       \affaddr{901 Marshal St.}\\
       \affaddr{Redwood City, CA}\\
       \email{bzhou@turn.com}
\alignauthor
Shahriar Shariat\\
       \affaddr{Turn Inc.}\\
       \affaddr{901 Marshal St.}\\
       \affaddr{Redwood City, CA}\\
       \email{sshariat@turn.com}
}
\date{23 November 2015}

\maketitle
\begin{abstract}
Online media offers opportunities to marketers to deliver brand messages to a large audience. Advertising technology platforms enables the advertisers to find the proper group of audiences and deliver ad impressions to them in real time. The recent growth of the real time bidding has posed a significant challenge on monitoring such a complicated system. With so many components we need a reliable system that detects the possible changes in the system and alerts the engineering team. In this paper we describe the mechanism that we invented for recovering the representative metrics and detecting the change in their behavior. We show that this mechanism is able to detect the possible problems in time by describing some incident cases.

\end{abstract}

%
%


%
%

%
%



\section{Introduction}
\label{sec:intro}
Online advertising has enabled the marketers to engage with the proper audiences in real time. This has resulted in exponential growth of online advertising and, in particular, real time bidding \cite{Yuan14}. The real time bidding platforms are complicated systems with many moving parts. Such platforms, typically, are highly distributed with low latency requirements. They offer variety of targeting and biding options and have diverse set of clients with radically different requirements, goals and budgets. The simple implication of such a heterogeneous environment is that, the likelihood of failure is not negligible. In other words, there is always a module, third party data or incoming traffic that can change and result in an abrupt and unexpected behavior of the system that can affect the entire set of advertisers, or just a group of advertising campaigns.

In order to monitor the health of the system, we, at Turn, have thousands of metrics that are monitored by various mechanisms. Many of such monitoring systems, rely on some anomaly detection algorithm. There has been a significant research work on anomaly detection. Refer to \cite{Chandola09,Chandola12} for comprehensive surveys on many anomaly detection algorithms and their applications. More recent works, such as \cite{Laptev15}, mainly focus on post processing techniques that reduce the false alarms and increase the quality of an anomaly detection task.Depending on the nature of the metric and its limitations, such as availability of labels, historical data, etc. , one needs to decide to choose from variety of algorithms and processing techniques.

 An important class of metrics is the final performance indicators. In advertising, these indicators include campaign's spending, number of submitted ad impressions, click through rate (CTR), conversion rate (CVR), etc. These metrics are the most interesting indicators for the advertisers. However, they are difficult to monitor since they are dependent on the incoming traffic, system performance and even the campaign setup, such as budget, audience targeting criteria, etc. 

We pose the monitoring of the final campaign performance metrics as a \emph{contextual anomaly detection} problem \cite{Chandola09}. We show that by clustering the relevant campaigns and selecting an appropriate representative for each cluster and other computational techniques, one can use a standard, and yet robust, anomaly detection algorithm to achieve a superior monitoring system. The contributions of this papers are:
\begin{itemize}
\item Enumerating and explaining the requirements and challenges of imposing a monitoring system on campaign performance metrics.
\item Proposing a method, based on correlation and statistical testing for representing the clusters of similar campaigns.
\item Presenting a comprehensive system for real time monitoring and alerting on final campaign performance metrics.
\end{itemize}

The rest of the paper is organized as follows. In \S~\ref{sec:background} we present some background on demand side platforms and its monitoring requirements. The detail of the system design and methodology is presented in \S~\ref{sec:design}. The evaluation of the the proposed system is presented in \S~\ref{sec:eval} and the paper is concluded by \S~\ref{sec:conclusion}.

\section{Background}
\label{sec:background}
A demand side platform (DSP), manages marketing campaigns of several advertisers simultaneously by identifying the target audience matching the campaign requirements and making real time bidding decisions on behalf of the advertisers to purchase ad impressions at the `best' price. Therefore, many advertisers have been compelled to use DSPs to deliver their branding messages to the right audience at scale. On the other hand, more users around the world are using the Internet to access the information on daily basis. Consequently, more advertising budgets together with more incoming traffic, have posed a great challenge on DSPs to scale and maintain the quality of service. In this section, we briefly describe the relevant background on the general design of a DSP and the overall monitoring system to set the stage for our proposed framework.
\subsection{Demand Side Platform}
\begin{figure}
\centering
\includegraphics[scale=.5]{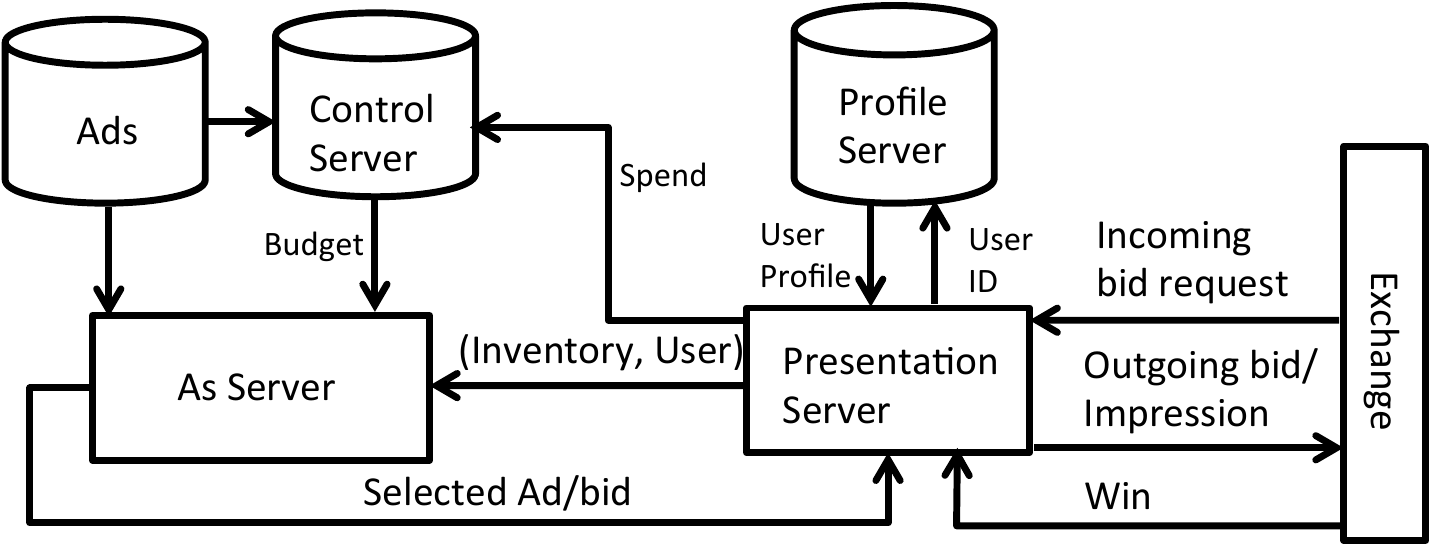}
\caption{The schematic design of a typical demand side platform.}
\label{lbl:DSP-schema}
\end{figure}
Figure \ref{lbl:DSP-schema} depicts the overall architecture of a demand side platform. The exchange (or the supply side platform) submits a bid request that describes the inventory by specifying the URL, position of the ad, acceptable ad format, ect., and, if available, a user ID. The request is received by the \emph{presentation server}, where it is decoded and augmented by a user profile that is extracted from a profile server, based on the provided user ID. This decoded message, sometimes called an \emph{adCall}, is then submitted to an \emph{ad server}. Ad server, receives the adCall, the set of ads and their budget, and determines the set of ads that are qualified to bid for the inventory. For each qualified ad, a bid is estimated that quantifies the value of the inventory to the advertiser for that specific ad. For instance, if the advertiser value a click at 2\$ and the the bid estimation algorithm estimates that the probability of the click is 1\%, then the bid price for this ad, inventory and user is 0.02\$. Refer to~\cite{Lee12,Shariat15}, for more detail on bid prediction and model evaluation.

After producing a bid for each one of the qualified ads, the ad server submits back the ad that has the highest estimated bid price, to the presentation server, which, after encoding, sends this information back to the exchange. The ad exchange platform, conduct an auction to identify the winner ad submitted by multiple DPS. If the submitted ad wins the auction, the DSP will be notified to serve the corresponding ad impression. This entire process needs to be completed within at most 100 millisecond including the network delay.

An important component in the above work flow, is a service that we call the \emph{control server}. Control server maintains the real time budget of each campaign, the number of impressions, the spending, clicks and conversion attributed to each campaign. The main duty of the control server is to allocate~\cite{Geyik14} and pace~\cite{Xu15} the budget of each campaign through its lifetime. Therefore, in contrast to other services, the control server is a singleton. The ad serving services, namely presentation, profile and ad servers, are highly distributed services. Therefore, a centralized service, i.e. the control server, is required to gather the information, specially, about the spending and remaining budget. This property gives the control server a unique position as the source of truth. Therefore, our anomaly detection mechanism is deployed in the control server. 

\subsection{Monitoring}
Having a highly complicated and distributed architecture, a DSP is prone to falter at any time. There are generally two sources of failure: i)internal and ii) external.

\textbf{Internal}: A bug in the most recent release is probably the most common internal issue. Despite comprehensive testing, it is still possible that a faulty code is deployed in the production. This can cause immediate crashes in the system or, it might cause only a set of campaigns to perform slightly worse. 

\textbf{External}: A DSP is dependent on its exchange partners. If they fail to provide a robust and reliable traffic, the performance of a DSP can be greatly affected. Also, many advertisers would like to use third-party or first-party data. For instance, an advertiser might need to use its own marketing data for targeting the correct audience. In that case, the DSP needs to ingest that data and use them during the ad serving. Another example is when an advertiser wants to use a mechanism offered by a third-party provider for filtering the web sites that its ads are served to. A failure on part of a data provider can cause the system to behave abnormally, despite the fact that all internal systems are performing as expected.

There is also, more sensitive services, such as fraud detection for which, one might want to use a combination of external and internal solutions.

One needs to impose monitoring systems on all the relevant metrics from incoming traffic to messages passed between the services and the internal health of each service such as its memory and CPU usage (per machine). It is evident that there are thousands of metrics that need to be monitored. 

A set of important metrics that represent the health of system and needs to be watched closely is the final output of the system. That is, the ad serving and the possible outcomes, i.e. clicks and conversions. Actually, as long as the campaigns are serving ads and spend their budget and observe clicks and conversions at a \emph{stable} pace, there is no urgent problem. Essentially, if for instance, an exchange partner fails but majority of the campaigns are able to switch their ad serving to another exchange, the urgency of the issue is much less than a failure that stops some campaigns from spending their budget. 

Therefore, our focus in this paper is to impose a reliable and robust monitoring system on the final output of the system, which is campaign spending and performance. 

\subsection{Goals}
The goal of our anomaly detection system is to issue an alert about once the DSP is not functioning as expected, in a timely manner so the engineers could investigate and fix the issues before considerable damage is caused.
\begin{itemize}
	\item The metric itself, must be as informative as possible. That is, once the alert is initiated, we must have a quick way to isolate the cause of malfunctioning.
    \item The alert should have a reasonably high precision. As we mentioned earlier, the alerts that are generated on the final campaign performance metrics are considered as a high priority and thus, it is important that we employ a reliable and precise mechanism for initiating them. We do not want to miss an incident either and therefore, we require the detection algorithm to posses a reasonable discovery rate as well. In fact we would prefer a to have a few false alarms in the interest of a higher discovery rate. 
	\item The alerts should be triggered within reasonable amount of time, typically a couple of hours since the onset of an incident.
\end{itemize}
 The latency requirement poses a challenge on the control server. To attain a holistic view of the campaign performance, the anomaly detection system collects data from all campaigns, which imposes overhead to several latency-sensitive core components in the DSP, such as budget pacing and data collection. As a result we need to minimize the performance impact by running the detection system less frequently and hence tolerate longer failure-to-detection latencies.

\input{design}

\input{eval}

%

\input{acmTargetAd.bbl}
%
%

\end{document}

%% file: design.tex
\section{Design}
\label{sec:design}

Millions of metrics are being generated constantly for all the campaigns running on Turn's platform, every one of which needs to be monitored and analyzed in real time to identify anomalous behaviors since system issues can occur at any part of the platform. The overhead could easily bring the platform to its knees especially so considering the sub-100-millisecond latency requirement imposed on bidding by ad exchanges. In this section, we present the design of our anomaly detection system which sieves through millions of metrics efficiently to find abnormal behaviors within a reasonable amount of time. The key idea of our design is to apply filtering and aggregation to the metrics before running the anomaly detection algorithm against the resulting data of a much smaller amount.

\begin{figure}
	\centering
	\includegraphics[width=\linewidth]{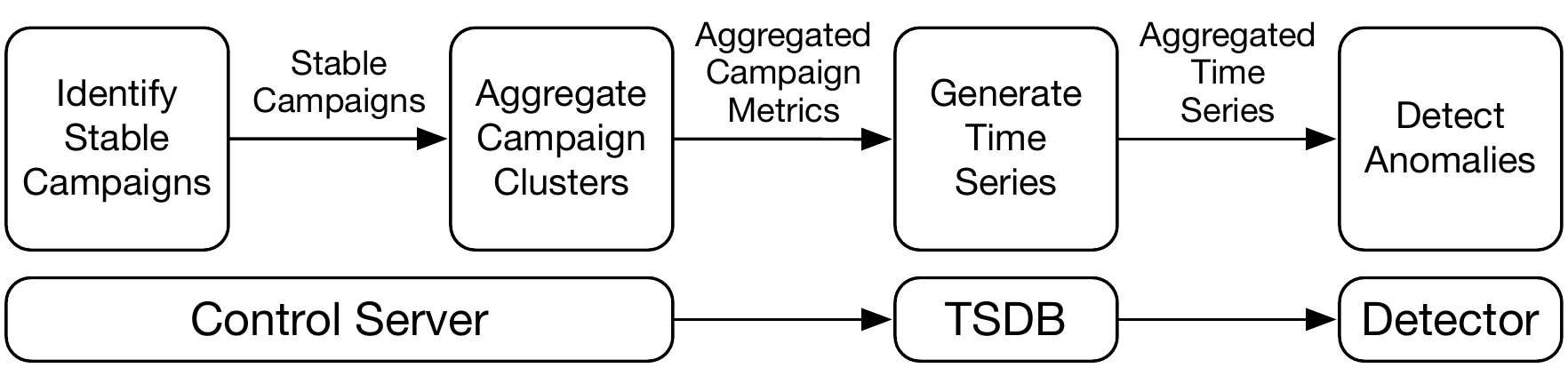}
	\caption{The Campaign Monitoring and Anomaly Detection Workflow.}\label{fig:workflow}
\end{figure}

As illustrated by Figure~\ref{fig:workflow}, the workflow of the anomaly detection system consists of four steps: (a) identifying stable campaigns by looking at campaign setup and behavior correlation; (b) aggregating the metrics over clusters of stable campaigns; (c) generating and storing time series from the aggregated metrics; (d) detecting anomalies using a simple algorithm designed to identify under-performing metrics. As described in Section~\ref{lbl:DSP-schema}, the control server, as a centralized service that maintains all campaign performance metrics, is the most suitable place to implement the anomaly detection system. The first two steps, stable campaign identification and metric aggregation are running on the control server. Time series data is submitted from a dedicated reporter thread running on the control server and stored in a distributed time series database. Finally, the anomaly detector is deployed as a separate service that ingest time series from the database to find anomalies and alert for system issues.

\subsection{Stable Campaign Identification}

Let $S_{stable}$ denote the set of \emph{stable} campaings. Given a metric, we scan all campaigns periodically and select the appropriate ones to include in (or exclude from) $S_{stable}$. Particularly, all campaigns are included $S_{stable}$ in the beginning, which is then prune through a two-step process detailed in the section. The selection of stable campaigns are refreshed every hour and cached in the memory by the control server.

\begin{table}
	\caption{Stability Checks for Campaign Setup}
	\label{tab:checks}
	\centering
	\begin{tabular}{|l|l|}
		\hline
		\textbf{Name} & \textbf{Requirement} \\
		\hline
		\hline
		Currency & US Dollar \\
		\hline
		Status & Active \\
		\hline
		Duration & > $p$ Days\\
		\hline
	\end{tabular}
\end{table}

\subsubsection{Campaign Setup}

During the first step, the configurations of all campaigns are checked against a set of pre-defined rules presented Table~\ref{tab:checks}. The campaigns that fail any of these tests in Table~\ref{tab:checks} are removed from $S_{stable}$. The check list is determined primarily based on our experience in tuning numerous campaigns. 

The anomaly detection system considers only US dollar based campaigns because the campaigns the number of campaigns that run on this currency represent a sufficiently large portion of the system. Furthermore, mixing the currencies will add to the already significant noise that the system carries. Note that the intention of this system is to catch system-wide failures and not campaign specific problems.
Furthermore, we only consider active campaigns to monitor the current status of the platform. Therefore the campaigns paused by advertisers or stopped due to budget exhaustion are also excluded from $S_{stable}$. On the other hand, we also avoid the campaigns started a short period ago since there is not much data for us to analyze when the campaigns are still in the optimization mode.



\subsubsection{Traffic Seasonality}

\begin{figure}
	\centering
	\includegraphics[width=.6\linewidth]{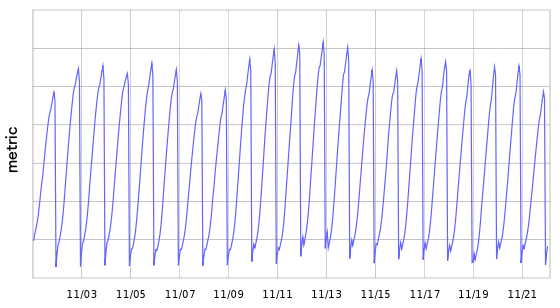}
	\caption{Seasonal Patterns in a Performance Metric.}
	\label{fig:seasonl}
\end{figure}

Now $S_{stable}$ contains only campaigns that have passed all the stability checks for campaign setup mentioned above. However, a stable campaign setup itself does not guarantee a stable performance since the incoming traffic qualified for a given campaign can vary over the time and make the campaign performance difficult to predict. Figure~\ref{fig:seasonl} illustrates the seasonal behaviors over the hour of the day and the day of the week of a performance metric due to traffic seasonality. 


To exclude the impact from traffic fluctuation, we compare the behavior of a campaign between two moving windows $w_1$ and $w_2$, each of $l$ hours long, dispersed by a $p$-day period, to determine if the campaign preserves the same behavior over $w_1$ and $w_2$ albeit the traffic patterns. Other things being equal, $w_1$ and $w_2$ should behavior similarly if an $l$-hour-long pattern repeats every $p$ days in the volume of the traffic. For the sake of simplicity, we apply $l=24$ and $p=1,7$ in our system to capture daily and weekly patterns in the traffic. The proper values for $l$ and $p$ can be learned from the time series itself using statistical modeling techniques, which we leave for future exploration.

\begin{figure}
	\centering
	\includegraphics[width=.45\linewidth]{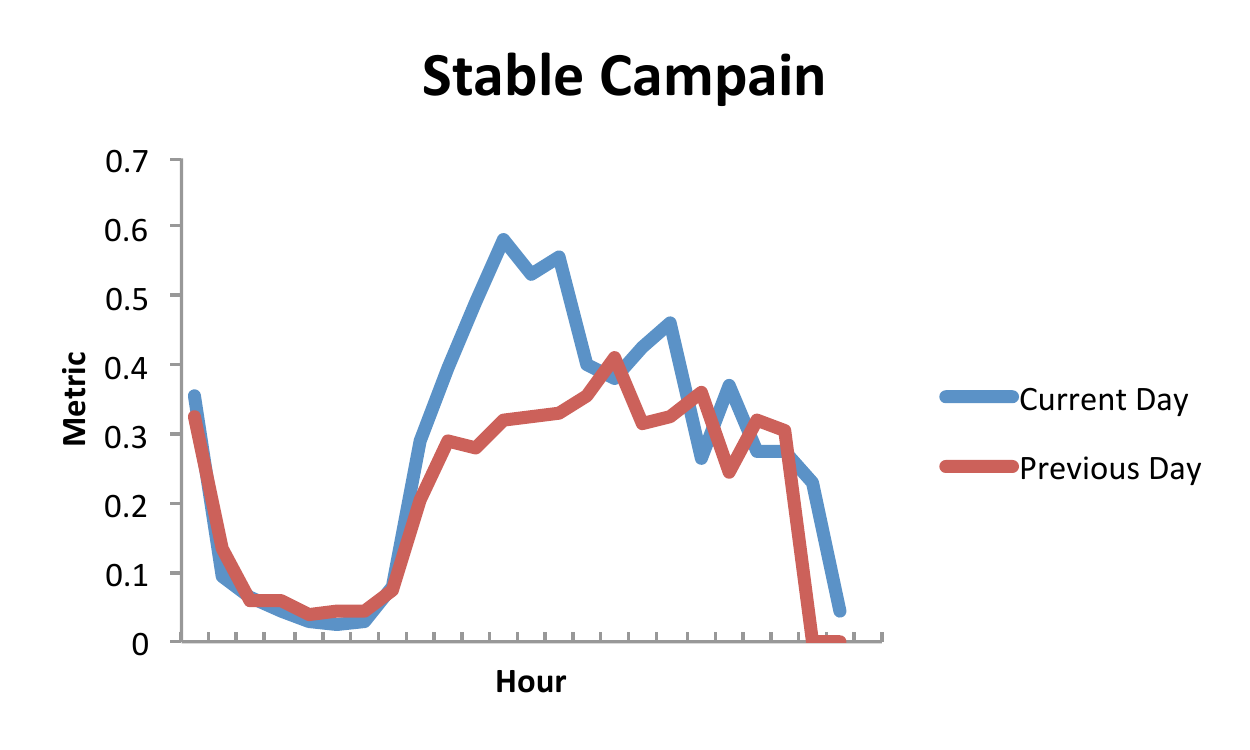}
	\includegraphics[width=.45\linewidth]{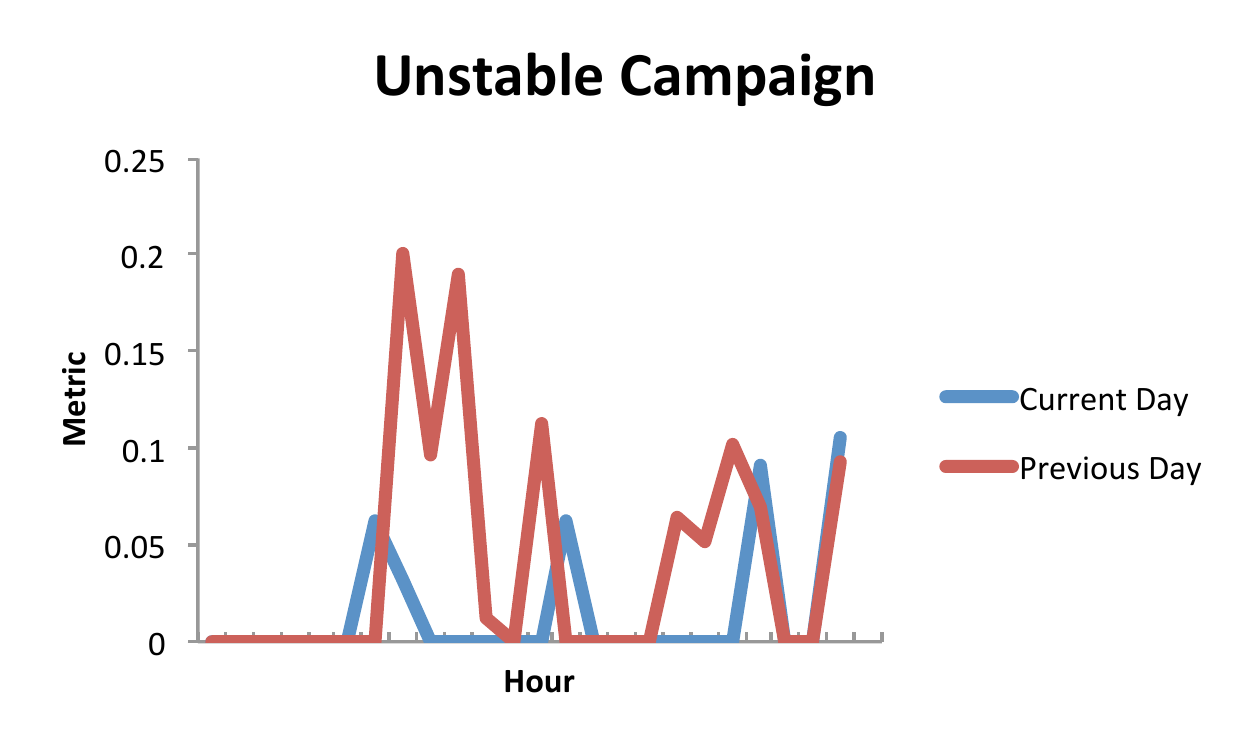}
	\caption{Identify stable campaigns using week-over-week behavior correlations.}\label{fig:stable}
\end{figure}

To quantify the behavior similarity between $w_1$ and $w_2$, we track the performance metric over the two time windows and discretize the curve into two vectors $v_1$ and $v_2$ that consist of hourly aggregated values for $w_1$ and $w_2$, respectively. Let $C_{w_1, w_2, l, p}$ denote the correlation of $v_1$ and $v_2$. We use $C_{w_1, w_2, l, p}$ as the measure of behavior similarity between $w_1$ and $w_2$. Furthermore, campaigns in $S_{stable}$ is filtered using $C_{w_1, w_2, l, p}$ and a constant $0< \delta < 1$ by this inequality:
\begin{align}
C_{w_1, w_2, l, p} > \delta
\end{align}

Since the current hour may exhibit anomalous behavior due to unforeseeable system issues, we introduce a delay of $x$ hours between the end hour of $w_1, w_2$ and the current hour. Essentially, as long as a campaign performs stably till $x$ hours ago, the campaign would be considered to have a correlated behavior by our system.


\subsection{Change Metric Generation}
\label{sec:metric}

Starting from the set of stable campaigns $S_{stable}$, we split the stable campaigns into clusters and aggregate the raw metrics in each cluster. Metric aggregation is performed over selected clusters of campaigns such that the anomaly detection system can cover different parts of the platform through monitoring the aggregated metrics. Campaigns are clustered according to the criteria used for targeted advertising or the media channels where the ads are served. Because our code base is constructed in a way that each targeting criterion or media channel is handled by a relatively self-sufficient module, we can easily pinpoint an issue in the platform according to the aggregated metrics that manifest the issue. The anomaly detection system considers the most common targeting criteria as listed in Table~\ref{tab:targeting} and four major media channels, namely display, video, mobile and social networks when clustering the campaigns.

\begin{table}
	\caption{Metric Aggregation based on Targeting Criteria of Campaigns}
	\label{tab:targeting}
	\centering
	\begin{tabular}{|l|p{5cm}|}
		\hline
		\textbf{Name} & \textbf{Target}\\
		\hline
		\hline
		Demographic & Users of particular demographic traits, such as age and gender.\\
		\hline
		Contextual & Content category, page quality, brand safety, etc.\\
		\hline
		Behavioral & Users that satisfy certain behavioral rules, e.g. having visited the advertiser's site before.\\
		\hline
		Dayparting & Particular time periods in a day.\\
		\hline
		Device & Particular types of devices, such PC, iOS, Android, etc.\\
		\hline
		Site List & Including or excluding a list of sites.\\
		\hline
	\end{tabular}
\end{table}



Downsampling is then performed on the aggregated time series to reduce to one data point per hour by summing up all data points within every hour. In doing so we preserve sufficient information for detecting anomalous behaviors within a couple of hours while avoiding the overhead that would incur if the original time series were transferred and analyzed. The sampled time series is transformed into a differential series by taking the difference of the metric between each pair of data points whose timestamps differ by $p$ days. We call the hourly differential time series \emph{change metric} to distinguish from the raw performance metric. The change metric is the final form of the data which is then stored in the time series database.

\subsection{Time Series Storage}

Time series are generated on the control server and then submitted to OpenTSDB~\cite{opentsdb}, a distributed time series database built on top of HBase~\cite{hbase}. OpenTSDB serves the persistence layer for the time series data and the feed for the anomaly detector.

Each sample in the time series consists of the timestamp when the entry is submitted, the value of the metric and a list of tags. The timestamps are written with second resolution. In case of duplicated data points for the same timestamp, the most recent value is recorded. Tags are a set of optional key-value pairs associated with the time series to provide additional information about the aggregation. Samples of the time series are taken periodically by the dedicated reporter thread on the control server and submitted to OpenTSDB through HTTP messages. A message begins with the operation name, followed by the metric name, the timestamp and the metric value, finally ends with the list of tags, as illustrated by the example below:
\begin{lstlisting}
   put proc.net.tcp.connections 1417642359 2 remote_host=50.116.234.5 direction=in state=established domain=sjc2 host=app454
\end{lstlisting}
The reporter thread collects data for the lightweight metrics in real time and reads the resource-intensive metrics from a cache refreshed less frequently by the control server to prevent performance degradation in the mission-critical parts of the control server, such as budget enforcement which deals with money.

OpenTSDB is deployed in a dedicated cluster to provide a scalable and reliable store for the enormous amount of time series data generated by thousands of machines in our production system. Upon the arrival of each sample, it performs compression to generate more compacted data that is faster to query and more suitable for dashboards and graphs spanning more than a few hours of data. OpenTSDB provides a convenient user interface for querying and visualizing the stored time series. It also supplies a set of HTTP APIs that enable querying, inserting and deleting the data in batches. Our anomaly detector is one of the consumers of the time series data stored in the OpenTSDB cluster and employs the batch APIs to fetch the data set to analyze.

\subsection{Anomaly Detection}

In this subsection, we describe our standard anomaly detection algorithm applied onto the change metric defined in Section~\ref{sec:metric}. The algorithm is presented in Table \ref{alg:anomaly-detect}. We assume a Gaussian distribution, $N(\mu, \sigma)$, on the input time series, that is $d_t, \forall t$. Essentially, the algorithm denotes a $d_i$ as anomalous if $|d_i-\mu| > \beta \sigma$ where $0\leq \beta \leq 3$ is a parameter that depends on the discovered anomalies. To make the algorithms more robust, one needs to limit its memory. To accomplish this, we use a decaying factor, namely $0 \leq \alpha \leq 1$, to smoothly discount the past during the calculation of running sum of time points. Therefore, the algorithm can easily be applied to any time-series and requires us to keep only three variables in the memory.
\begin{algorithm}
	\caption{Anomaly detection algorithm}
	\begin{algorithmic}
		\STATE{\textbf{Input}}
		\STATE{$d_t$}
		\STATE{\textbf{Initialization}}\\
		\STATE{$\mathcal{X},\mathcal{X}^2,n$}
		\WHILE{$d_t$ is valid}
		\STATE{$\mu = \frac{\mathcal{X}}{n}$,$\sigma=\sqrt{\frac{\mathcal{X}^2}{n}-\left(\frac{\mathcal{X}}{n}\right)^2}$}
	
		\IF{$|d_t-\mu| > \beta \sigma$}
		\IF{$d_t < 0$}
		\STATE{$a_t = Anomaly$}
		\STATE{$shrink(\beta)$}
		\ENDIF
		\ELSE
		\STATE{$\mathcal{X}= \alpha \mathcal{X} + d_t$}
		\STATE{$\mathcal{X}^2=\alpha \mathcal{X}^2+d_t^2$}
		\STATE{$n = \alpha n + 1$}
		\ENDIF
		\ENDWHILE
	\end{algorithmic}
	\label{alg:anomaly-detect}
\end{algorithm}

The initialization phase of the algorithm takes a small initial part of the time series as the training data and estimates the values for $\mathcal{X}, \mathcal{X}^2, n$ based on the training data which is assumed anomaly-free. In the main loop of the algorithm, $\mathcal{X}, \mathcal{X}^2, n$ are updated for every data points falling within the $\mu \pm \beta \sigma$ range. Note only the negative outliers are deemed true anomalies and decrease the range of normal data points by shrinking $\beta$, the rest including both normal data points and positive outliers are considered normal by the algorithm. The focus of our work is on detecting anomalies when they indicate a negative change. While a positive change can be anomalous as well, they are controlled and detected by other systems. For instance, if we start receiving abnormally large number of clicks, then our fraud detection modules start to investigate the issue. Therefore, our system follows an asymmetric  design that focuses on \emph{under-performing}. Also, a sudden jump is system's performance metrics is expected when an issue is resolved.
We use $\beta$ as the shrinkage for the normal range and diminish the range proportionally according to the percentage of anomalies detected in a moving window. Denote $N_{normal}$ the number of normal points in the window and $N_{abnormal}$ the number of anomalies. Then $\beta$ is updated in function $shrink(\beta)$ as follows:


\begin{align}
\beta = 3 \cdot \frac{N_{normal}}{N_{normal} + N_{abnormal}}
\end{align}

%% file: eval.tex
\section{Evaluation}
\label{sec:eval}

\begin{figure*}[tbhp]
	\includegraphics[width=\linewidth]{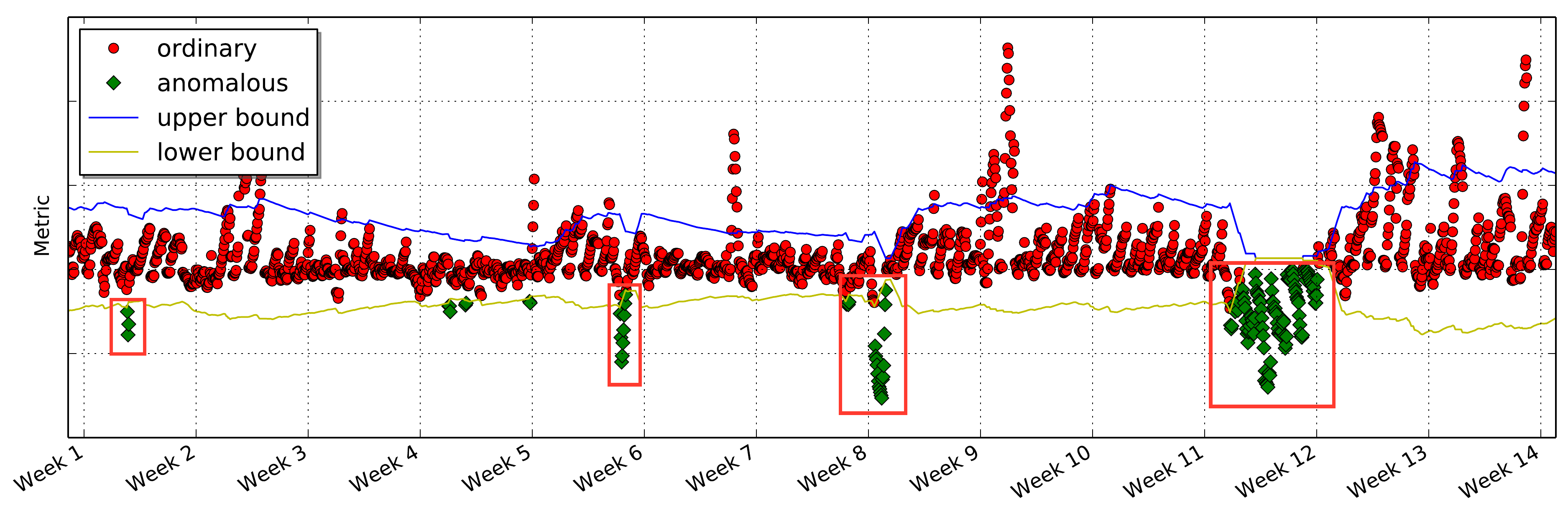}
	\caption{The anomaly detection algorithm is applied to a real performance metric collected over three months. \textcolor{red}{Red} boxes highlight the periods in which the incidents occurred. \textcolor{green}{Green} diamonds are the data points flagged as anomalous by our tool. The upper and lower bounds are determined by $\mu \pm \beta \sigma$ in Algorithm~\ref{alg:anomaly-detect}.}
	\label{fig:anomaly-detection}
	\vspace{-0.5cm}
\end{figure*}
This section presents our experience applying the anomaly detection system to a real performance metric during a period in which four incidents were discovered by the tool. Two of the issues were transient and resolved by themselves. The other two issues were caused by program errors. Three of the incidents lasted for less than a day, while the last one got fixed a week after its inception. Our tool successfully detected all four issues and demonstrated its robustness against both short-living and long-lasting issues. The final change metric retrieved by the anomaly detector consists of 2185 data points, out of which 160 points are anomalous as they concur with system issues in time. One week worth of data is used as the training data to initialize the anomaly detection algorithm and the rest of the data is used to evaluate the performance of the system. Note we omit the numbers and exact timestamps associated with the metrics in the results presented here to comply with the company's regulations.

\subsection{Metric Stability}

\begin{figure}[tbhp]
	\centering
	\subfigure[A day-over-day metric that includes all campaigns.]{%
		\includegraphics[width=\linewidth]{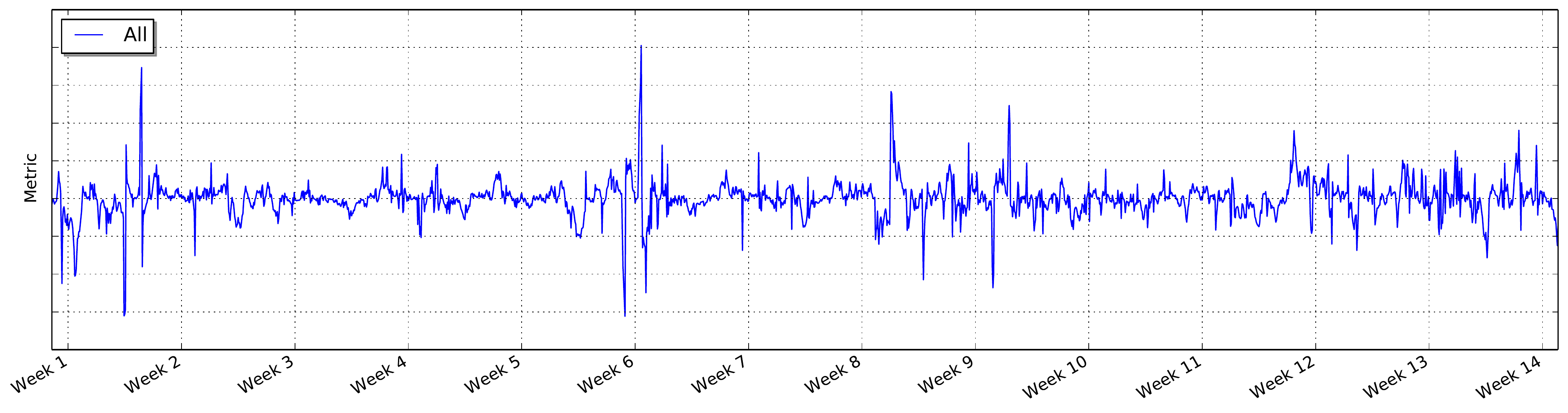}\label{fig:compare-daily-all}%
	}
	\subfigure[Day-over-day change metrics that include only stable campaigns.]{%
		\includegraphics[width=\linewidth]{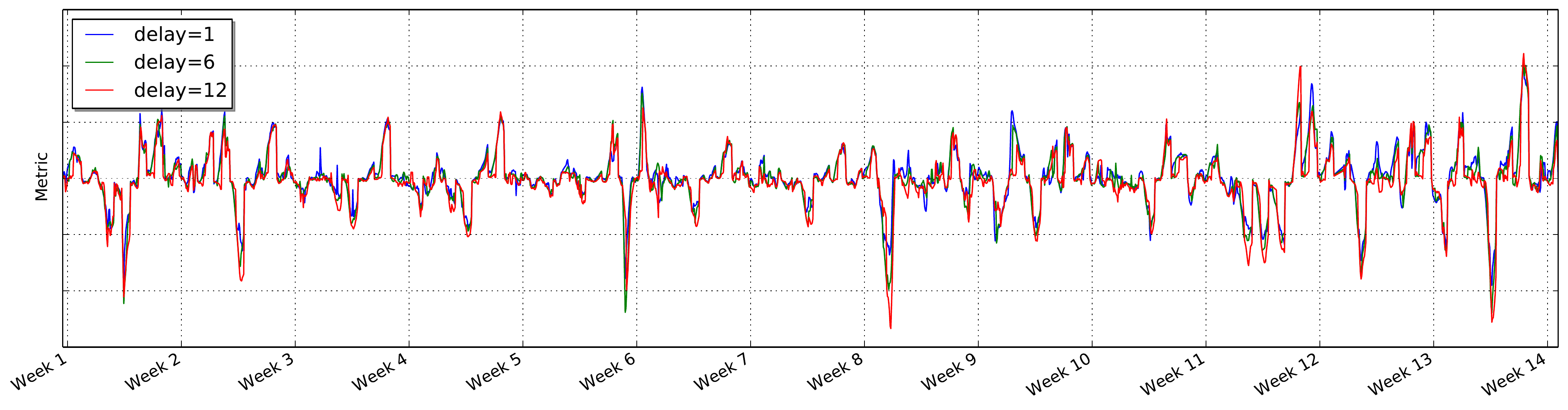}\label{fig:compare-daily-metric}%
	}
	\caption{The daily metrics are unstable in general due to traffic fluctuation during the course of the week, and difficult to correlate with the occurrence of system issues.}
\end{figure}

\begin{figure}[tbhp]
	\centering
	\subfigure[A week-over-week metric that includes all campaigns.]{%
		\includegraphics[width=\linewidth]{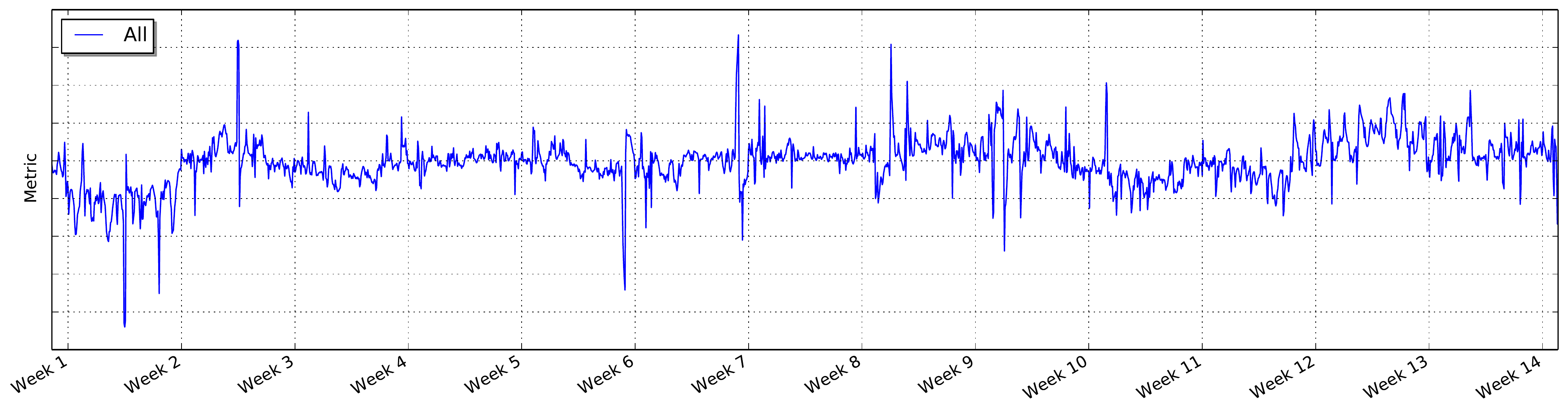}\label{fig:compare-weekly-all}%
	}
	\subfigure[Week-over-week change metrics that include only stable campaigns.]{%
		\includegraphics[width=\linewidth]{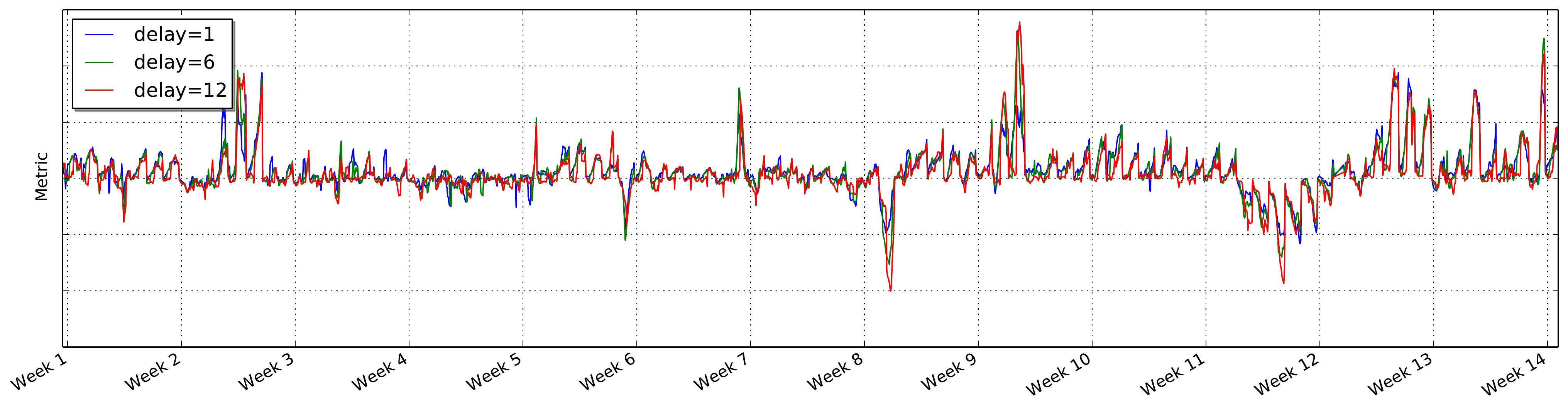}\label{fig:compare-weekly-metric}%
	}
	\caption{The weekly metrics from stable campaigns are highly correlated with prior system issues.}
\end{figure}

We compare the metrics obtained from all campaigns running on Turn's platform and those including only stable campaigns. The day-over-day change metrics (i.e. $p=1$) in Figure~\ref{fig:compare-daily-metric} is evidently more informative than the all-campaign metric in Figure~\ref{fig:compare-daily-all} regarding the occurrences of system issues. The week-over-week change metrics (i.e. $p=7$) in Figure~\ref{fig:compare-weekly-metric} shows a highly stabilized behavior which leaves the negative spikes caused by system issues apparent to identify. On the contrary, the week-over-week all-campaign metric in Figure~\ref{fig:compare-weekly-all} reveals little information about system health as it is contaminated by strong noise. In our final configuration of the anomaly detection system, we use the week-over-week metric in Figure~\ref{fig:compare-weekly-metric} as the input to the detector.

\subsection{Accuracy}

We also evaluate the accuracy of the anomaly detection system using the same data set described above. The week-over-week change metric derived from the original time series is used for the numbers presented here. We employ the standard $F_1$ measure defined in Equation~\ref{eq:f1} to quantify the accuracy of the anomaly detection system. In Equation~\ref{eq:f1}, \emph{precision} denotes the ratio of true anomalies among all detected data points, and \emph{recall} the percentage of anomalies that are detected.
\begin{align}
\var{F_1-Score} = 2 \cdot \frac{Precision \cdot Recall}{Precision + Recall}\label{eq:f1}
\end{align}
Figure~\ref{fig:anomaly-detection} illustrates the result of the experiment: 157 true anomalies and 12 false positives are identified and 3 anomalies missed. The system has a precision of 92.9\% and a recall of 98.1\%. The $F_1$ measure is 0.954 based on the precision and the recall. The false positives are mostly borderline cases which overlay or slightly exceed the lower bound for anomaly detection, while all of the missed anomalies locate close to the beginning of incidents when it is difficult to discern between normal and abnormal behaviors based on metrics.

\section{Conclusion}
\label{sec:conclusion}

We present an anomaly detection system that identifies system issues based on self-correlated behavior of stable campaigns in a large-scale DSP. The system has been deployed in our production cluster and successfully detected several major system problems which otherwise would not be discovered until severe consequences are caused.